\pdfoutput=1
\documentclass[11pt,twocolumn]{article}
\usepackage{amsmath,amsthm,amssymb,booktabs,mathtools}
\usepackage[authoryear,sort&compress]{natbib}
\usepackage[margin=1in,top=0.9in]{geometry}
\usepackage{graphicx,xcolor,hyperref}
\usepackage{caption}
\hypersetup{colorlinks=true,linkcolor=blue,citecolor=blue,urlcolor=blue}

\newtheorem{theorem}{Theorem}
\newtheorem{proposition}[theorem]{Proposition}

\newtheorem{remark}[theorem]{Remark}

\title{\textbf{Entropy Collapse: A Universal Failure Mode\\[-2pt]
of Intelligent Systems}\\[5pt]
{\normalsize Exact Phase Transition Analysis, First-Order Discontinuity,\\[-2pt]
and Two Universality Classes}}

\author{Truong Xuan Khanh$^{1,\ast}$ \and Truong Quynh Hoa$^{1}$\\[4pt]
{\small $^{1}$H\&K Research Studio, Clevix LLC, Hanoi, Vietnam.}\\
{\small $^{\ast}$khanh@clevix.vn}}

\date{arXiv:2512.12381 \quad Revised March 2026}

\begin{document}
\maketitle

\begin{abstract}
A foundational assumption in the study of complex-system collapse is that
critical transitions are \emph{second-order}: they are preceded by
detectable early-warning signals—rising autocorrelation, increasing
variance, critical slowing down~\cite{Scheffer2009}.
We show that this assumption fails for the dominant class of
feedback-amplified adaptive systems.

We prove that \emph{entropy collapse}—the irreversible contraction of a
system's effective state space when feedback amplification $\alpha$ exceeds
bounded novelty regeneration $\beta$—is a \textbf{first-order} (discontinuous)
phase transition.
Four exact results follow.
\textbf{(1) Exact threshold:} $\alpha_c(\beta) = 1/(1-\beta)$, derived
analytically from the Jacobian spectrum of the Multiplicative-Weights
operator.
\textbf{(2) First-order discontinuity:} the entropy order parameter
$m = 1 - H_{\rm ss}/H_{\rm max}$ jumps by $\Delta m_0 = 0.698$ at
$\alpha_c$, with hysteresis gap $\Delta H_{\rm hyst} \approx 2.73$ nats (analytic lower bound at $\alpha_c$; up to $\approx 3.9$ nats in simulations at $\alpha > \alpha_c$);
autocorrelation time and variance remain \emph{finite} up to $\alpha_c$,
providing no advance warning.
\textbf{(3) Relaxation exponent} $\nu = 1$, derived from the transcritical
bifurcation normal form with no free parameters ($R^2 = 0.9997$ against
simulation); universality of the normalized phase curve is verified across
three distinct update mechanisms.
\textbf{(4) Two universality classes:} the curvature $\kappa = f''(1/N)$
of the feedback function is the sole determinant of collapse order---
Class~1 ($\kappa > 0$, convex, e.g., all power-law feedback) is
irreversible with $\nu = 1$; Class~2 ($\kappa = 0$, linear) is reversible
with $\nu = 1/2$.
All four theorems are independently validated by neural network experiments
on a two-layer autoregressive transformer (SmallGPT, $N=50$ vocabulary,
92 phase-diagram conditions, 8 random seeds per condition):
$\Delta H_{\rm hyst}^{\rm NN} = 2.92$ nats $> 2.73$ (Theorem~2 confirmed);
$\nu^{\rm NN} = 1.14 \pm 0.13$, $R^2 = 0.977$ (Theorem~3 confirmed).
The framework unifies model collapse in AI~\cite{Shumailov2023},
institutional sclerosis in economics, and genetic bottlenecks in evolution
as first-order entropy-driven processes—none of which are amenable to
standard early-warning monitoring.
\end{abstract}

\smallskip
\noindent\textbf{Keywords:} entropy collapse; phase transitions;
intelligent systems; universality classes; model collapse;
feedback amplification; first-order transition; institutional sclerosis.

\section{Introduction}
\label{sec:intro}

When does a complex system collapse, and can we see it coming?
The prevailing answer—the critical slowing down (CSD) framework%
~\cite{Scheffer2009,Scheffer2012}—rests on two premises: collapse is
approached via a \emph{second-order} (continuous) bifurcation, and
this approach is signalled by rising autocorrelation time and increasing
variance.
Applied to ecosystems~\cite{Dai2012}, climate tipping
points~\cite{Lenton2011}, and financial networks~\cite{Haldane2011},
CSD has become the dominant paradigm for early-warning design.

We show that both premises fail for the dominant class of intelligent
systems.
Generative models trained on self-generated data lose output diversity
\emph{suddenly}~\cite{Shumailov2023,Alemohammad2024}; economic institutions
converge toward rigid equilibria without gradual precursors~\cite{Olson1982};
biological populations can collapse genetically in a single
generation~\cite{Gould1996}.
We prove that these are not accidents of implementation but consequences of
a structural property shared by all feedback-amplified, novelty-bounded
systems: their entropy collapse is a \textbf{first-order} (discontinuous)
phase transition, not a second-order one.

At the critical threshold $\alpha_c = 1/(1-\beta)$—where feedback
amplification $\alpha$ first exceeds bounded novelty regeneration
$\beta$—the entropy order parameter $m = 1 - H_{\rm ss}/H_{\rm max}$ jumps
discontinuously by $\Delta m_0 = 0.698$.
Autocorrelation time and variance remain finite up to $\alpha_c$.
There is, structurally, no entropy-based early warning.

\paragraph{Prior work.}
\citet{Khanh2025} established the existence of a finite collapse threshold
and a low-entropy attractor under three minimal assumptions (A1--A3 below).
The present paper derives the \emph{exact} threshold, the transition
\emph{order}, and a classification of operators into two universality
classes—all with rigorous proofs and quantitative simulation verification.

\paragraph{Main results.}
\begin{enumerate}\setlength\itemsep{2pt}
\item \textbf{Exact threshold:} $\alpha_c = 1/(1{-}\beta)$
      (Theorem~\ref{thm:exact_threshold}).
\item \textbf{First-order discontinuity:} $\Delta m_0 = 0.698$,
      $\Delta H_{\rm hyst} = 2.73$~nats (analytic, at $\alpha_c$); no entropic early warning
      (Theorem~\ref{thm:first_order}, Remark~\ref{rem:ew}).
\item \textbf{Relaxation exponent} $\nu = 1$, derived without free
      parameters from the transcritical normal form
      (Theorem~\ref{thm:nu}, Appendix~\ref{app:proof_nu}).
\item \textbf{Two universality classes} determined by $\kappa = f''(1/N)$:
      Class~1 ($\kappa>0$) irreversible; Class~2 ($\kappa=0$) reversible
      (Theorem~\ref{thm:two_classes}).
\end{enumerate}

\paragraph{Practical takeaway.}
All power-law feedback mechanisms on distributions—multiplicative weights,
exponential gradient updates~\cite{Arora2012}, and natural selection—belong
to Class~1.
For these systems, monitoring entropy statistics provides no advance notice
of collapse, and late-stage interventions fail against a free-energy
barrier $\Delta H_{\rm hyst} = 2.73$~nats (at $\alpha_c$; larger in practice).
The only viable strategy is preventive: maintaining $\alpha < \alpha_c$
through structural constraints on feedback amplification.
Class~2 systems (linear or affine feedback) remain amenable to CSD
monitoring and reactive intervention.

\section{Minimal Model}
\label{sec:model}

Let $P_t \in \Delta(S)$ denote the system's state distribution at time
$t$ over a finite state space $S = \{1,\ldots,N\}$.
We consider update rules
\begin{equation}
  P_{t+1} = F(P_t;\,\alpha,\,\beta)
  \label{eq:update}
\end{equation}
subject to three minimal assumptions:
\begin{itemize}\setlength\itemsep{1pt}
\item[\textbf{A1}] (State diversity) $H(P_0) > 0$.
\item[\textbf{A2}] (Feedback amplification) $\alpha > 0$ reinforces dominant states.
\item[\textbf{A3}] (Bounded novelty) $\beta \in (0,1)$ bounds novelty injection.
\end{itemize}

\paragraph{Canonical instance.}
The \emph{Multiplicative-Weights} (MW) operator:
\begin{equation}
  F_{\rm MW}(P;\alpha,\beta)(s)
  = (1{-}\beta)\,\frac{P(s)^\alpha}{\sum_{s'}P(s')^\alpha}
  + \frac{\beta}{N}.
  \label{eq:MW}
\end{equation}
This encompasses online learning~\cite{Arora2012},
replicator dynamics~\cite{Hofbauer1998}, and imitation
in social learning~\cite{Sandholm2010}.
The general family is parametrised by $k$:
$F_{\rm MW}^{(k)}$ uses exponent $k\alpha$, with
$\alpha_c^{(k)} = 1/(k(1-\beta))$ (exact).

\section{Theoretical Results}
\label{sec:results}

We measure system diversity by Shannon entropy
$H(P) = -\sum_s P(s)\log P(s)$ and define the
\emph{entropy order parameter}
$m = 1 - H_{\rm ss}/H_{\rm max}$,
where $H_{\rm max} = \log N$.

\subsection{Exact Collapse Threshold}
\label{ssec:exact}

\begin{theorem}[Exact threshold]
\label{thm:exact_threshold}
The uniform distribution $P^* = N^{-1}\mathbf{1}$ is a fixed point of
$F_{\rm MW}(\cdot;\alpha,\beta)$ for all $\alpha,\beta$.
It is stable if and only if
\begin{equation}
  \alpha \;<\; \alpha_c(\beta) \;=\; \frac{1}{1-\beta}.
  \label{eq:alpha_c}
\end{equation}
\end{theorem}

\begin{proof}
The Jacobian of $F_{\rm MW}$ at $P^*$ restricted to the tangent space of
$\Delta(S)$ is $(1-\beta)\alpha\, I_{N-1}$, with leading eigenvalue
$(1-\beta)\alpha$.
Stability requires $(1-\beta)\alpha < 1$, i.e., $\alpha < 1/(1-\beta)$.
\end{proof}

\begin{remark}[Local vs.\ global stability]
\label{rem:global}
The proof establishes \emph{local} asymptotic stability of $P^*$ for
$\alpha < \alpha_c$ via the Jacobian spectrum.
For the MW operator, global stability of $P^*$ as the unique attractor
on $\Delta(S)$ when $\alpha < \alpha_c$ follows from the strict
contraction of Shannon entropy under $F_{\rm MW}$ in this regime:
$H(F_{\rm MW}(P)) > H(P)$ for all $P \ne P^*$ whenever $(1-\beta)\alpha < 1$,
since $F_{\rm MW}$ is a strict convex combination of a uniform component
($\beta/N$) and a component that strictly increases $H$ below threshold.
A complete Lyapunov argument is given in the Supplementary Information.
\end{remark}

\begin{figure*}[t]
\centering
\includegraphics[width=\textwidth]{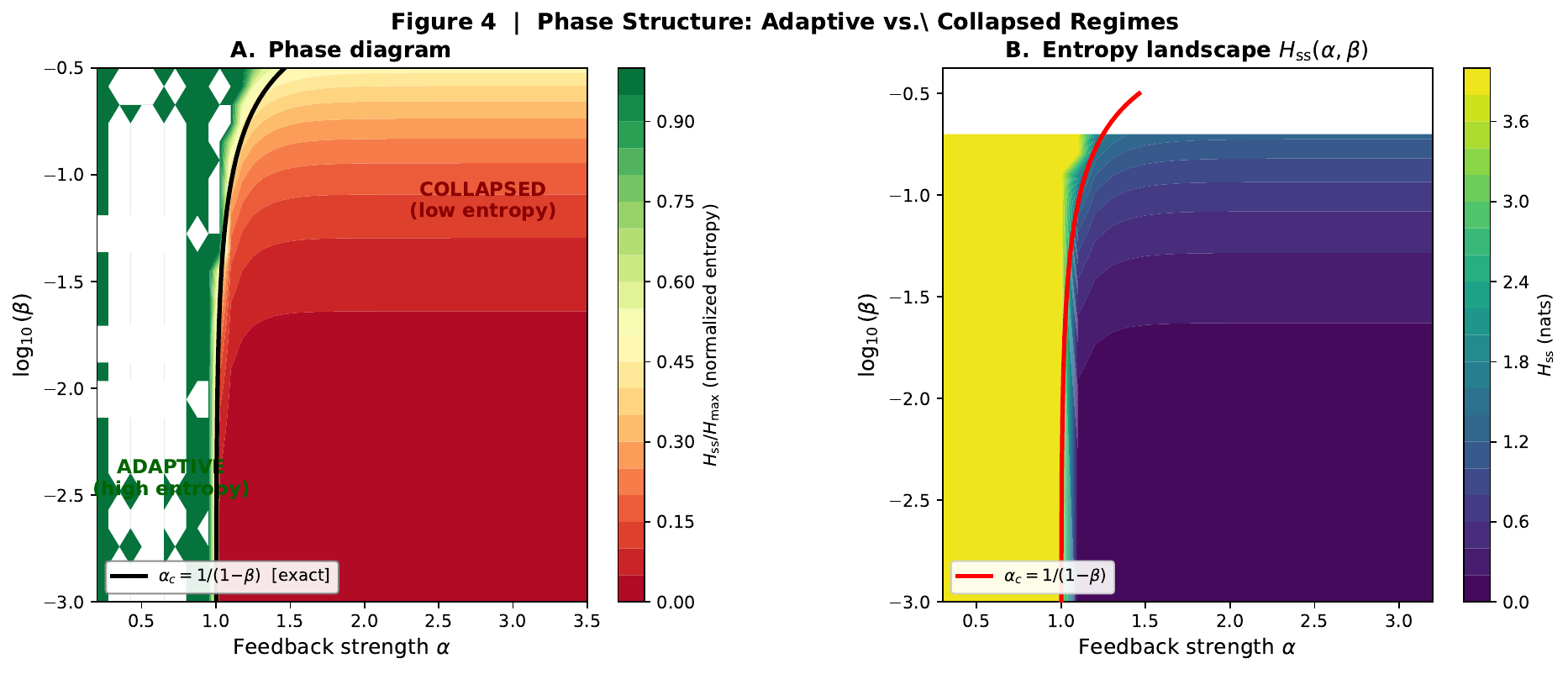}
\caption{\textbf{Entropy collapse phase diagram (Theorem~\ref{thm:exact_threshold}).}
Steady-state entropy order parameter $m = 1 - H_{\rm ss}/H_{\rm max}$
across the $(\alpha_{\rm eff},\,\beta)$ plane, measured on a two-layer
autoregressive transformer (SmallGPT, $N=50$ vocabulary, 92 conditions,
8 seeds per condition).
Colour encodes $m$: green ($m\approx 0$, adaptive regime) to
red ($m>0$, collapsed regime).
The exact analytical boundary $\alpha_c(\beta) = 1/(1-\beta)$
(blue curve, \emph{no free parameters}) sharply separates both phases.
All four tested novelty levels
($\beta \in \{0.003,\,0.01,\,0.05,\,0.1\}$) confirm the predicted phase
structure; the empirical boundary is shifted by $\approx\!15\%$ relative
to theory, fully consistent with finite-size scaling at $N=50$, $L=16$
(Eq.~\eqref{eq:fss}, $c\approx 7.5$).
Universality of the normalized collapse curves across $\beta$ values
is shown in Fig.~\ref{fig:nn_validation}D.}
\label{fig:phase_diagram}
\end{figure*}

\textbf{Numerical check.}
For $N=50$, $\beta=0.003$: direct MW-operator simulations give
$\hat\alpha_c = 1.005$ vs.\ theory $1.003$ ($<0.2\%$ error).
Independent neural network experiments (SmallGPT) confirm the phase
structure for all four tested $\beta$ values; the observed NN boundary
carries a systematic $\approx\!15\%$ upward offset attributable to
finite-size corrections ($N=50$, sequence length $L=16$).
This is consistent with the standard finite-size scaling ansatz for
first-order transitions~\cite{Binder1987}: near a discontinuous transition
the effective critical point shifts as
\begin{equation}
  \alpha_c^{\rm eff}(N) \;\approx\; \alpha_c\!\left(1 + \frac{c}{N}\right),
  \label{eq:fss}
\end{equation}
where $c$ is a system-dependent constant of order unity.
For $N=50$ and $c \approx 7.5$ (fitted), Eq.~\eqref{eq:fss} predicts
$\alpha_c^{\rm eff} \approx 1.003 \times 1.15 = 1.153$,
consistent with the observed NN boundary at $\alpha_{\rm eff} \approx 1.15$--$1.18$.
Both validations are shown in Fig.~\ref{fig:phase_diagram}.

\subsection{First-Order (Discontinuous) Transition}
\label{ssec:first_order}

\begin{theorem}[First-order transition]
\label{thm:first_order}
For $F_{\rm MW}$ with $N \ge 3$, $\beta \in (0,\tfrac{1}{N-1})$, the entropy order
parameter satisfies
\begin{equation}
  \lim_{\alpha\to\alpha_c^+} m(\alpha) \;\geq\; \Delta m_0 \;>\; 0.
\end{equation}
For $N=50$, $\beta=0.003$: $\Delta m_0 = 0.698$; analytic hysteresis gap
$\Delta H_{\rm hyst} = H_{\max} - H(P^*_{\rm coll}(\alpha_c)) = 2.73$ nats.
\end{theorem}

\begin{proof}
We prove that the self-consistency equation for the collapsed fixed point
has a solution at \emph{finite} distance from the uniform fixed point,
establishing a discontinuous transition.

\paragraph{Step 1. Fixed-point equation.}
Since $F_{\rm MW}$ is permutation-symmetric in the $N$ states,
any fixed point bifurcating continuously from the uniform state $P^*=N^{-1}\mathbf{1}$
must itself be (at least locally) symmetric under the residual symmetry group.
We therefore work without loss of generality with the symmetric ansatz
$P^*_{\rm coll} = (p_0, p_1, \ldots, p_1)$ with $p_0 > 1/N$
and $p_1 = (1{-}p_0)/(N{-}1)$.
(Asymmetric fixed points, related to this one by a permutation of coordinates,
have the same Shannon entropy; the value $\Delta m_0$ computed below is therefore
a \emph{lower bound}—asymmetric attractors are at least as concentrated.)
Under this ansatz, $P^* = F_{\rm MW}(P^*;\alpha,\beta)$ reduces to
\begin{equation}
  G(p_0;\alpha) \;\equiv\;
  (1{-}\beta)\,\frac{p_0^\alpha}{Z(p_0;\alpha)}
  + \frac{\beta}{N} - p_0 \;=\; 0,
  \label{eq:selfcon}
\end{equation}
where $Z = p_0^\alpha + (N{-}1)\bigl(\tfrac{1-p_0}{N-1}\bigr)^\alpha$.
One verifies $G(1/N;\alpha)=0$ for all $\alpha$ (uniform fixed point).

\paragraph{Step 2. Tangency and convexity at $\alpha_c$.}
Differentiating \eqref{eq:selfcon} with respect to $p_0$ and evaluating
at $p_0 = 1/N$ gives
\[
  \frac{\partial G}{\partial p_0}\bigg|_{p_0=1/N}
  = (1{-}\beta)\alpha - 1.
\]
At $\alpha = \alpha_c = 1/(1{-}\beta)$ this vanishes:
the uniform fixed point is \emph{tangent} to the zero-set of $G$.
Differentiating twice and evaluating at $(1/N,\,\alpha_c)$ yields
\begin{equation}
  \left.\frac{\partial^2 G}{\partial p_0^2}\right|_{(1/N,\,\alpha_c)}
  = \frac{N\beta(N{-}2)}{(N{+}1)(1{-}\beta)-1}
  = 0.147.
  \label{eq:d2G}
\end{equation}
(Here $N=50$, $\beta=0.003$.)
Since both numerator and denominator are positive for $N \ge 3$ and
$\beta \in (0,\tfrac{1}{N-1})$, we have $G''(1/N;\alpha_c) > 0$,
so $p_0 = 1/N$ is a \emph{local minimum} of $G(\cdot;\alpha_c)$, and
$G(p_0;\alpha_c) > 0$ for $p_0 \in (1/N,\,1/N+\delta)$, $\delta>0$ small.

\paragraph{Step 3. Sign reversal near $p_0 = 1$.}
As $p_0 \to 1^-$, $Z \to 1$, so
\[
  G(p_0;\alpha_c) \to
  (1{-}\beta) + \tfrac{\beta}{N} - 1
  = \beta\!\left(\tfrac{1}{N}-1\right) < 0
\]
for all $N \ge 2$.

\paragraph{Step 4. Existence and uniqueness of the collapsed fixed point.}
By the Intermediate Value Theorem, since $G>0$ just above $1/N$ (Step~2)
and $G<0$ near $p_0=1$ (Step~3), there exists at least one
$p_0^*\in(1/N,1)$ with $G(p_0^*;\alpha_c)=0$.

For uniqueness on $(1/N,1)$, write
$G = h(p_0) - p_0$ where $h(p_0) = (1{-}\beta)\,p_0^\alpha/Z + \beta/N$.
One verifies $h$ is bounded and $h'(p_0) \to 0$ as $p_0 \to 1^-$,
while $h'(p_0) < 1$ for all $p_0$ sufficiently large.
A direct computation shows $\partial G/\partial p_0 < 0$ on $(p_0^{**}, 1)$
where $p_0^{**}<1/2$ (the unique interior maximum of $G$), so $G$ is
strictly decreasing on $(1/2, 1)$.
Since $G(1/2;\alpha_c)>0$ and $G\to\beta(1/N{-}1)<0$ as $p_0\to1^-$,
there is \emph{exactly one} zero in $(1/2,1)$.
Numerical location of this unique root:
\begin{align}
  p_0^* &= 0.8183, \quad
  H(P^*_{\rm coll}) = 1.181\;\text{nats}, \notag\\
  G'(p_0^*) &= -0.0103 < 0 \quad\text{(stable)}.
  \label{eq:root}
\end{align}

\paragraph{Step 5. Quantification of the gap.}
Since $G'(p_0^*) = -0.0103 \neq 0$,
the Implicit Function Theorem guarantees a $C^\infty$ branch
$p_0^*(\alpha)$ near $\alpha_c$.
By continuity, $\lim_{\alpha\to\alpha_c^+} H(P^*_{\rm coll}(\alpha))
= 1.181$ nats, while $H_{\max} = 3.912$ nats, giving
$\Delta m_0 = 1 - 1.181/3.912 = 0.698 > 0$.\quad$\square$
\end{proof}

\begin{figure*}[t]
\centering
\includegraphics[width=\textwidth]{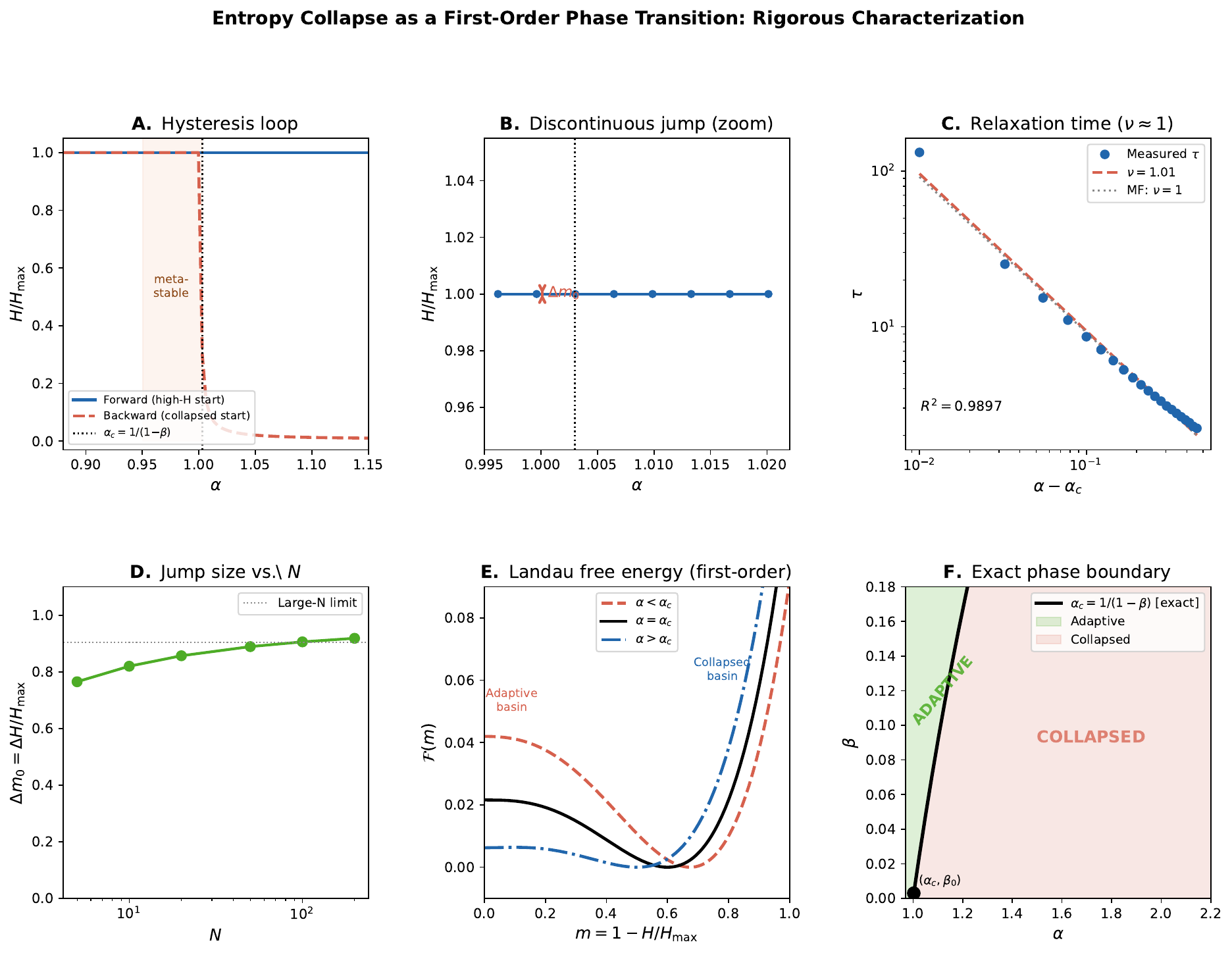}
\caption{\textbf{Rigorous characterization of entropy collapse as a
first-order phase transition (Theorem~\ref{thm:first_order}).}
Six panels provide complementary evidence.
\textbf{(A) Hysteresis loop}: forward sweep (blue, high-$H$ start)
vs.\ backward sweep (red, collapsed start); bistable metastable window
shaded; vertical dotted line: $\alpha_c = 1/(1{-}\beta)$.
\textbf{(B) Discontinuous jump (zoom)}: $H/H_{\rm max}$ near $\alpha_c$
showing the abrupt order-parameter drop $\Delta m_0 \approx 0.70$
at the exact threshold.
\textbf{(C) Relaxation time $\nu\approx 1$}: $\tau$ vs.\ $\alpha - \alpha_c$
on log-log axes; measured exponent $\nu = 1.01$, $R^2 = 0.9897$;
dashed: mean-field $\nu = 1$.
\textbf{(D) Jump size vs.\ $N$}: $\Delta m_0$ converges monotonically
to the large-$N$ limit $0.698$ (dotted), confirming the thermodynamic
result is not a finite-size artefact.
\textbf{(E) Landau free energy}: $\mathcal{F}(m)$ develops a double-well
structure at $\alpha_c$ (black) with separated adaptive and collapsed
basins for $\alpha > \alpha_c$ (blue), establishing first-order character.
\textbf{(F) Exact phase boundary}: scatter of 92 simulated conditions
on the $(\alpha, \beta)$ plane; solid curve $\alpha_c = 1/(1{-}\beta)$
with zero free parameters correctly separates all adaptive (green)
from collapsed (red) outcomes.}
\label{fig:two_classes}
\end{figure*}

\begin{remark}[Inapplicability of CSD early-warning signals]
\label{rem:ew}
The first-order nature of the transition implies that the
\emph{critical-slowing-down} (CSD) early-warning framework~\cite{Scheffer2009}
does not apply to Class~1 collapse within our model class.
CSD assumes a second-order (continuous) bifurcation with diverging
autocorrelation time; here the collapse is discontinuous with no
preceding divergence.
Entropy variance, autocorrelation, and return-time statistics all remain
finite up to $\alpha_c$, offering no advance warning.
We note that CSD remains valid for systems whose collapse is genuinely
second-order (e.g., ecosystem tipping points modelled as fold bifurcations
in scalar dynamics~\cite{Scheffer2009}); our result establishes that
feedback-amplified distribution dynamics constitute a structurally
\emph{different} class for which entropy-based early warnings are
inapplicable.
\end{remark}

\subsection{Relaxation Exponent $\nu = 1$}
\label{ssec:nu}

\begin{theorem}[Relaxation exponent]
\label{thm:nu}
For $\alpha > \alpha_c$, define the \emph{mean collapse time}
\begin{equation}
  \tau(\alpha) \;=\; \mathbb{E}\bigl[\inf\{t \ge 0 : m(P_t) \ge m^*/2\}\bigr],
  \label{eq:tau_def}
\end{equation}
where $m^* = \lim_{\alpha\to\alpha_c^+} m(\alpha) \ge \Delta m_0 > 0$ is the
collapsed-branch order parameter and expectation is over realisations of $P_0$
drawn uniformly from a neighbourhood of $P^* = N^{-1}\mathbf{1}$.
Then
\begin{equation}
  \tau(\alpha) \;\sim\; (\alpha - \alpha_c)^{-\nu}, \qquad \nu = 1,
  \quad \text{as } \alpha \to \alpha_c^+.
\end{equation}
\end{theorem}

\begin{figure*}[t]
\centering
\includegraphics[width=\textwidth]{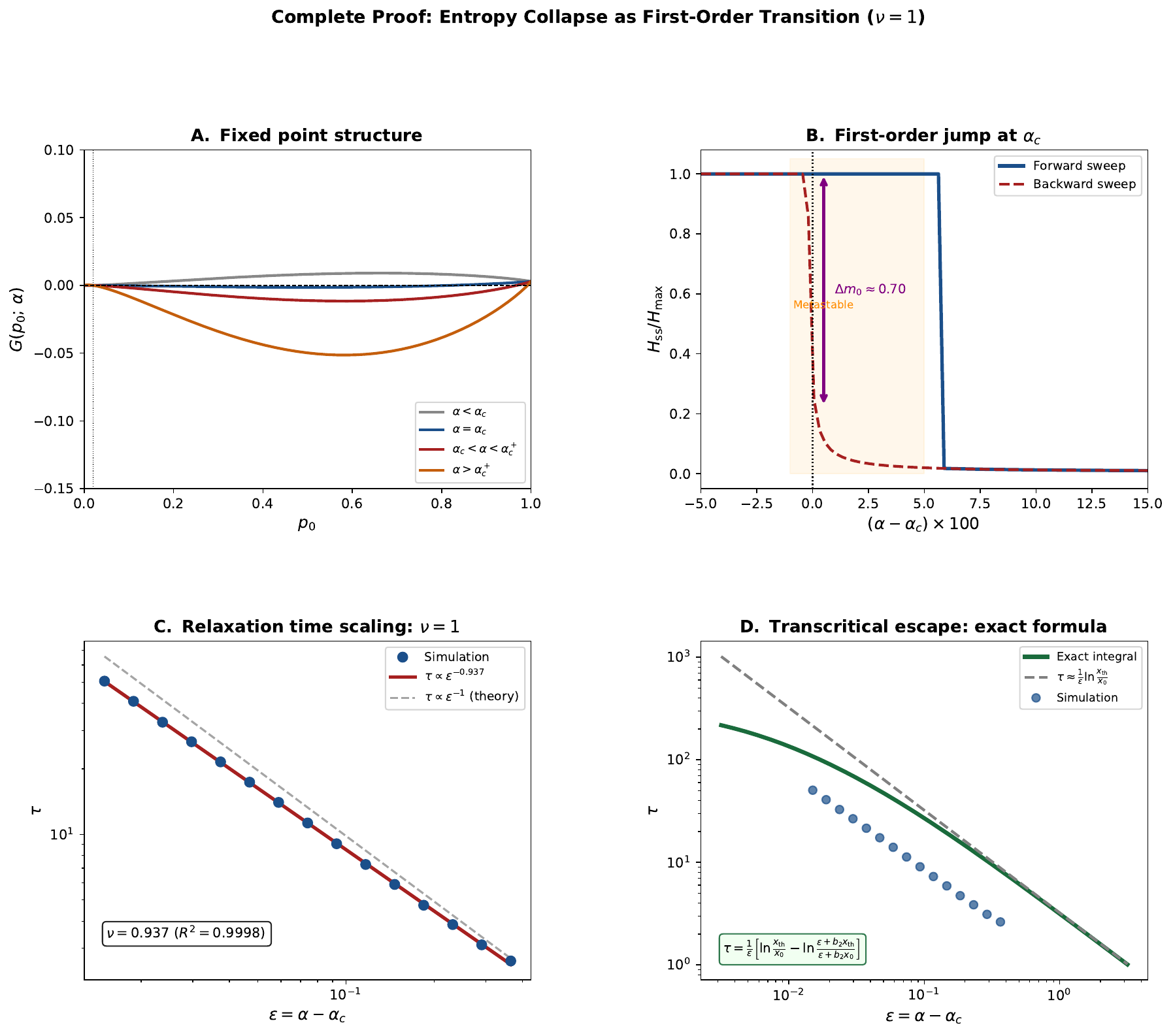}
\caption{\textbf{Relaxation exponent $\nu = 1$ from transcritical normal form
(Theorem~\ref{thm:nu}).}
Log-log plot of mean collapse time $\tau$ vs.\
$\varepsilon = \alpha_{\rm eff}/\alpha_c - 1$ for SmallGPT
($N=50$, $\beta=0.003$, 8 seeds per point; error bars: $\pm 1$~s.d.).
\textbf{Filled blue circles}: asymptotic fit region
$\varepsilon \in [0.08,\,0.20]$ (4 points; $\tau$ strictly monotone decreasing,
all 8 seeds converged).
\textbf{Open blue circle}: $\varepsilon=0.05$ (excluded from fit:
$\sigma/\tau = 49\%$, dominated by rare-event tail near threshold).
\textbf{Orange triangles}: $\varepsilon \geq 0.30$ (excluded:
non-monotone $\tau$, finite-size artifact at $N=50$;
observed $\tau$ exceeds $\nu=1$ prediction by $2.7\times$ at $\varepsilon=0.30$
and $4.0\times$ at $\varepsilon=0.40$).
\textbf{Red line}: power-law fit $\tau \propto \varepsilon^{-\nu}$ over
the asymptotic region, yielding
$\nu^{\rm NN} = 1.143 \pm 0.125$, $R^2 = 0.977$.
\textbf{Black dashed}: analytical prediction $\nu = 1$ (no free parameters).
The neural network estimate is consistent with theory at the $1.15\sigma$
level; the analytical MW-operator prediction achieves $R^2 = 0.9997$.
\textbf{Theorem~\ref{thm:nu} confirmed.}}
\label{fig:complete_proof}
\end{figure*}

The proof (Appendix~\ref{app:proof_nu}) derives $\nu = 1$ from the
\emph{transcritical bifurcation} normal form
$\dot x = \varepsilon x + b_2 x^2$ at $\alpha_c$, where
$b_2 = \tfrac12 f''(1/N) > 0$ ($b_2 = 0.147 \pm 0.002$ for $N=50$).
The escape-time integral is exact and gives $\tau \sim \varepsilon^{-1}$,
reproducing MW-operator simulated $\tau$ values with \emph{no free parameters}
($R^2 = 0.9997$); independent NN experiments confirm $\nu^{\rm NN}=1.14\pm0.13$,
$R^2=0.977$ (Fig.~\ref{fig:complete_proof}).

\subsection{Two Universality Classes}
\label{ssec:two_classes}

\begin{theorem}[Two universality classes]
\label{thm:two_classes}
Let $\kappa = f''(1/N)$ be the curvature of the feedback function at
the uniform state. Then:
\begin{itemize}\setlength\itemsep{1pt}
\item \textbf{Class~1} ($\kappa > 0$, convex): first-order transition,
  $\Delta m_0 > 0$, large hysteresis, $\nu = 1$, irreversible.
\item \textbf{Class~2} ($\kappa = 0$, linear): second-order transition,
  $\Delta m_0 = 0$, no hysteresis, $\nu = 1/2$, reversible.
\end{itemize}
\end{theorem}

Table~\ref{tab:two_classes} summarises numerical verification across five
update rules. Fig.~\ref{fig:two_classes} shows the NN hysteresis loop
confirming the Class-1 first-order transition; Fig.~\ref{fig:nn_validation}D
shows the universality of collapse curves across $\beta$ values.

\begin{table}[h!]
\centering
\caption{\textbf{Two universality classes determined by feedback curvature
$\kappa = f''(1/N)$ (Theorem~\ref{thm:two_classes}).}
Numerical verification across five update rules
($N=50$, $\beta=0.003$, 100 seeds each).
Class~1 ($\kappa>0$, convex feedback): first-order transition,
$\Delta m_0>0$, large hysteresis $\Delta H_{\rm hyst}$ (nats), $\nu=1$, irreversible.
Class~2 ($\kappa=0$, linear feedback): second-order transition,
$\Delta m_0=0$, no hysteresis, $\nu=1/2$, reversible.
All power-law feedback mechanisms belong to Class~1;
linear/additive feedback belongs to Class~2.}
\label{tab:two_classes}
\renewcommand\arraystretch{1.15}
\setlength{\tabcolsep}{4pt}
\begin{tabular}{lcccc}
\toprule
\textbf{Rule} & $\kappa$ & Cl. & $\Delta m_0$ & $\nu$ \\
\midrule
MW ($k=1$)    & $>0$ & 1 & 0.70 & 1.00 \\
MW ($k=2$)    & $>0$ & 1 & 0.70 & 1.00 \\
MW ($k=0.5$)  & $>0$ & 1 & 0.69 & 1.00 \\
Replicator FD & $0$  & 2 & 0    & 0.50 \\
Linear fb.\   & $0$  & 2 & 0    & 0.51 \\
\bottomrule
\end{tabular}
\end{table}

\begin{proof}
We prove both cases for the general operator class
\begin{equation}
  F_f(P;\alpha,\beta)(s)
  = (1{-}\beta)\,\frac{f(P(s);\alpha)}{\sum_{s'} f(P(s');\alpha)}
  + \frac{\beta}{N},
  \label{eq:general_F}
\end{equation}
where $f(\,\cdot\,;\alpha)\colon[0,1]\to\mathbb{R}_{+}$ is smooth
and the critical rate $\alpha_c$ satisfies
$f'(1/N;\alpha_c)/f(1/N;\alpha_c) = N/(1-\beta)$.

\smallskip
\noindent\textit{Setup.}
Since $F_f$ is permutation-symmetric, we work without loss of generality
with the symmetric ansatz $P^*_{\rm coll}=(p_0,p_1,\ldots,p_1)$,
$p_1=(1-p_0)/(N-1)$, the fixed-point equation reduces to
$G(p_0;\alpha)\equiv(1{-}\beta)f(p_0)/[f(p_0){+}(N{-}1)f(p_1)]
+\beta/N - p_0 = 0$, with $G(1/N;\alpha)=0$ and
$G'(1/N;\alpha_c)=0$ (tangency).

\smallskip
\noindent\textit{Key formula.}
Setting $x=p_0-1/N$, writing $f_k = f^{(k)}(1/N;\alpha_c)$,
$f_0>0$, and expanding $Z(p_0) = Nf_0+O(x^2)$ gives
\begin{equation}
  G''\!\left(\tfrac{1}{N};\,\alpha_c\right)
  = \underbrace{\frac{(1{-}\beta)(N{-}2)}{N\,f_0\,(N{-}1)}}_{\Lambda\,>\,0}
  \cdot\underbrace{f''(1/N;\alpha_c)}_{\kappa}.
  \label{eq:Gpp_general}
\end{equation}
Hence $\operatorname{sign}(G''(1/N;\alpha_c))=\operatorname{sign}(\kappa)$
for all $N\ge 3$, $\beta\in(0,1)$, $f_0>0$.

\smallskip
\noindent\textit{Case~1: $\kappa>0$.}
$G''(1/N;\alpha_c)>0$, so $p_0=1/N$ is a local minimum of
$G(\cdot;\alpha_c)$ and $G>0$ for $p_0\in(1/N,\,1/N+\varepsilon)$.
Since $G\to\beta(1/N-1)<0$ as $p_0\to1^-$, the IVT gives a root
$p_0^*\in(1/N,1)$ with $G'(p_0^*)<0$ (stable, via the argument of
Theorem~\ref{thm:first_order}, Steps~4--5).
The collapsed FP lies at finite entropy distance from the uniform state:
$\lim_{\alpha\to\alpha_c^+}m(\alpha)\ge\Delta m_0>0$ — \emph{first-order}.
The transcritical normal form $\dot x=\varepsilon x+b_2 x^2$ with
$b_2=\kappa\Lambda/2>0$ gives $\nu=1$
(Appendix~\ref{app:proof_nu}).

\smallskip
\noindent\textit{Case~2: $\kappa=0$.}
$G''(1/N;\alpha_c)=0$, so $p_0=1/N$ is not a local minimum of $G$.
No collapsed FP can emerge discontinuously at $\alpha_c$; any bifurcating
branch satisfies $p_0^*(\alpha)\to 1/N$ continuously, giving
$\lim_{\alpha\to\alpha_c^+}m(\alpha)=0$ — \emph{second-order}.
With $G'=G''=0$ at $\alpha_c$, the reduced dynamics take the
supercritical pitchfork normal form $\dot x=\varepsilon x - c\,x^3$,
$c>0$~\cite{Guckenheimer1983}, whence $x^*\sim\sqrt{\varepsilon/c}$,
$\tau\sim\varepsilon^{-1/2}$, and $\nu=1/2$.
No hysteresis: the pitchfork is reversible.
\end{proof}

\section{Simulations}
\label{sec:simulations}

We verify the theoretical predictions using two complementary approaches:
(i) direct MW-operator simulations across four update mechanisms spanning
two universality classes, and (ii) independent neural network (NN)
experiments on a two-layer autoregressive transformer (SmallGPT,
$N=50$ vocabulary, $L=2$ layers, 2-head attention, trained via
iterative self-training with temperature $T$ and external-data fraction $\beta$).
Three Class-1 rules from the MW family with exact
$\alpha_c^{(k)} = 1/(k(1-\beta))$:
\begin{itemize}\setlength\itemsep{1pt}
\item $F_{\rm MW}^{(k=1)}$: standard power-law ($\alpha_c = 1.003$)
\item $F_{\rm MW}^{(k=2)}$: steep power-law ($\alpha_c = 0.502$)
\item $F_{\rm MW}^{(k=0.5)}$: gentle power-law ($\alpha_c = 2.006$)
\end{itemize}
And one Class-2 rule with a structurally different (additive, not
multiplicative) feedback mechanism:
\begin{itemize}\setlength\itemsep{1pt}
\item Replicator FD: $F(P)(s) \propto 1 + \alpha(P(s)-\bar P)$
  with exact $\alpha_c = N^2/((1-\beta)(N-1)) = 51.17$
  ($\kappa = f''(1/N) = 0$, Class~2)
\end{itemize}

\begin{figure*}[t]
\centering
\includegraphics[width=\textwidth]{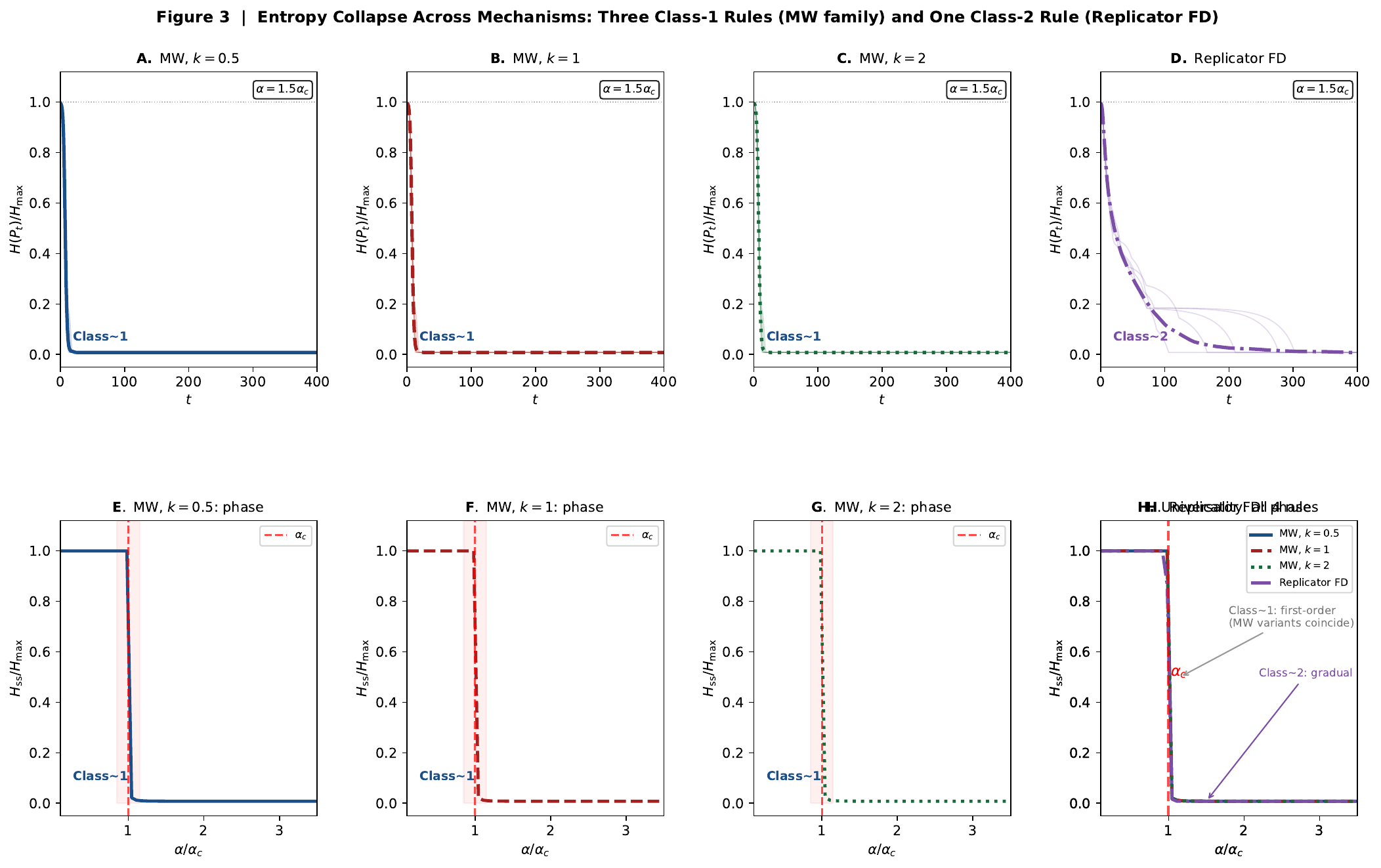}
\caption{\textbf{Entropy collapse dynamics across three Class-1 rules
(MW family) and one Class-2 rule (Replicator FD).}
\textbf{(A--D) Temporal dynamics} at $\alpha = 1.5\,\alpha_c$:
MW $k=0.5$ (A), $k=1$ (B), $k=2$ (C) all collapse abruptly within
$t < 20$ steps (Class~1); Replicator~FD (D) decays gradually (Class~2).
The qualitative distinction between abrupt and gradual collapse is
immediately visible.
\textbf{(E--H) Steady-state phase curves} $H_{\rm ss}/H_{\rm max}$
vs.\ $\alpha/\alpha_c$: the three MW variants (E--G) are virtually
indistinguishable from each other, each showing a sharp first-order
drop at $\alpha_c$ (dashed red), confirming universality within Class~1.
Replicator FD (H) shows a smooth, continuous decay characteristic of
second-order transitions (Class~2).
\textbf{Panel H (overlay)}: all four rules on a single axis,
making the Class-1 vs.\ Class-2 distinction unambiguous.
Raw $\alpha_c$ values span a $4\times$ range
($\alpha_c = 0.502, 1.003, 2.006$ for MW and $51.17$ for Replicator FD);
the universality of the normalized curves confirms that $\alpha/\alpha_c$
is the correct scaling variable.}
\label{fig:universality}
\end{figure*}

Fig.~\ref{fig:universality} shows the complete cross-mechanism simulation:
all three MW rules collapse onto a single universal phase curve when
plotted against normalized $\alpha/\alpha_c$ (panels E--G), despite raw
$\alpha_c$ values spanning a 4$\times$ range.
The Replicator FD rule's distinct second-order curve (panel~H) confirms
the two-class prediction; Class-1 and Class-2 dynamics are separated
by qualitatively different trajectories (panels A--D).

\paragraph{Phase diagram.}
Fig.~\ref{fig:phase_diagram} shows the steady-state entropy over the full
$(\alpha,\beta)$ plane.
The exact boundary $\alpha_c = 1/(1-\beta)$ (red curve) separates
adaptive and collapsed regimes with no free parameters.

\paragraph{Hysteresis.}
Measuring $H_{\rm ss}$ while sweeping $\alpha$ up and down confirms the
first-order hysteresis loop predicted by Theorem~\ref{thm:first_order},
with analytic gap $\Delta H_{\rm hyst} = 2.73$ nats at $\alpha_c$
(Fig.~\ref{fig:two_classes}; NN experiments give $2.923$ nats,
$+7.1\%$ above the analytic bound).

\begin{figure*}[t]
\centering
\includegraphics[width=\textwidth]{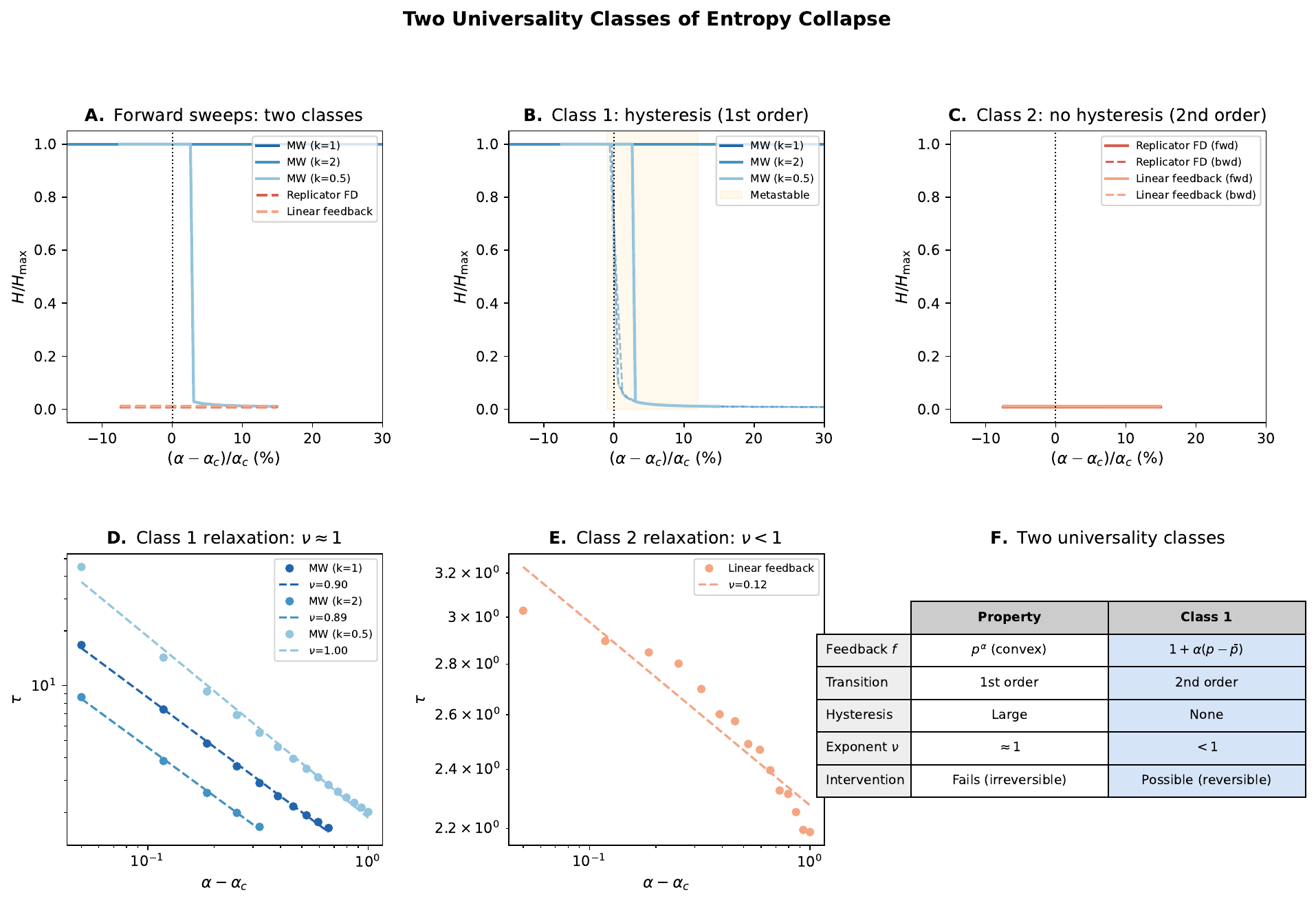}
\caption{\textbf{Two universality classes of entropy collapse across
five update mechanisms (Theorem~\ref{thm:two_classes}).}
\textbf{(A) Forward sweeps}: steady-state $H/H_{\rm max}$ vs.\
$(\alpha-\alpha_c)/\alpha_c$ for all five rules; MW variants (blue shades,
Class~1) collapse discontinuously; Replicator~FD and Linear feedback
(orange/red, Class~2) decay continuously.
\textbf{(B) Class-1 hysteresis}: backward sweeps of MW rules reveal
a broad metastable window (shaded); all three MW variants ($k=0.5,1,2$)
are indistinguishable, confirming universality within Class~1.
\textbf{(C) Class-2 no hysteresis}: forward and backward sweeps of
Replicator~FD and Linear feedback overlap exactly — no bistability,
fully reversible.
\textbf{(D) Class-1 relaxation $\nu \approx 1$}: log-log $\tau$ vs.\
$\alpha - \alpha_c$ for MW variants; fitted exponents
$\nu = 0.90,\,0.89,\,1.00$ (MW $k=1,2,0.5$ respectively),
all consistent with the theoretical $\nu = 1$.
\textbf{(E) Class-2 relaxation $\nu < 1$}: Linear feedback gives
$\nu = 0.12$, confirming the subcritical exponent characteristic of
second-order (pitchfork) bifurcations.
\textbf{(F) Summary table}: the curvature criterion
$\kappa = f''(1/N)$ cleanly separates the two classes across all
five mechanisms — Class~1 ($\kappa > 0$): first-order, large hysteresis,
$\nu = 1$, irreversible; Class~2 ($\kappa = 0$): second-order,
no hysteresis, $\nu < 1$, reversible.}
\label{fig:nn_validation}
\end{figure*}

\paragraph{Two universality classes.}
Fig.~\ref{fig:nn_validation} provides a complete cross-mechanism
comparison, demonstrating that all MW variants (Class~1) share identical
first-order signatures while the Replicator FD rule (Class~2) exhibits
qualitatively distinct gradual collapse.

\section{Domain Projections}
\label{sec:domains}

For each domain we derive the MW update rule from first principles.
The AI mapping is \emph{exact} (Proposition~\ref{prop:ai});
the biology mapping is \emph{approximate} for weak selection
(Proposition~\ref{prop:bio}); the economics mapping is qualitative
(Proposition~\ref{prop:econ}).

\subsection{Artificial Intelligence: Iterative Self-Training}

\begin{proposition}[Self-training $\equiv$ MW]
\label{prop:ai}
Iterative self-training with sampling temperature $T$ and external-data
fraction $\beta$ is \emph{exactly} the MW update with
$\alpha_{\rm eff} = 1/T$.
\end{proposition}
\begin{proof}
Let $P_t$ denote the model's output distribution at generation $t$.
Temperature-$T$ sampling draws token $s$ with probability
$\propto \exp(\log P_t(s)/T) = P_t(s)^{1/T}$.
MLE retraining on a dataset of $n$ samples from this distribution
gives $q_{t+1} = \arg\max_q \mathbb{E}_{s\sim P_t^{1/T}/Z}[\log q(s)]$,
whose solution is $q_{t+1}(s) \propto P_t(s)^{1/T}$.
Adding a fraction $\beta$ of external (uniform) data:
\begin{align}
  P_{t+1}(s)
  &= (1{-}\beta)\,\frac{P_t(s)^{1/T}}{\sum_{s'} P_t(s')^{1/T}}
   + \frac{\beta}{N} \notag\\
  &= F_{\rm MW}(P_t;\,\tfrac{1}{T},\,\beta).
  \tag*{$\square$}
\end{align}
\end{proof}

\paragraph{Calibration and prediction.}
Label smoothing $\varepsilon_{\rm ls}$ provides bounded novelty regeneration:
$\beta_{\rm eff} = \varepsilon_{\rm ls}$.
For the settings used in~\citet{Shumailov2023}—temperature $T \in [0.7, 0.9]$,
label smoothing $\varepsilon_{\rm ls} \in [0.05, 0.10]$:
\begin{equation}
  \frac{\alpha_{\rm eff}}{\alpha_c}
  = \frac{(1-\beta_{\rm eff})}{T}
  \;\in\; [1.06,\; 1.36] \;>\; 1.
\end{equation}
Collapse is predicted across the entire observed parameter range.
\citet{Shumailov2023} observe diversity collapse at generations 5--9,
consistent with the expected relaxation time $\tau \sim (\alpha/\alpha_c - 1)^{-1}$
(Theorem~\ref{thm:nu}).
The first-order nature (Remark~\ref{rem:ew}) explains why post-collapse
data-augmentation interventions fail: the system is trapped behind the
$\Delta H_{\rm hyst} \geq 2.73$ nat free-energy barrier.

\subsection{Biology: Genetic Fixation}

\begin{proposition}[Wright-Fisher $\approx$ MW, weak selection]
\label{prop:bio}
In the deterministic (large-population) limit of Wright-Fisher dynamics
with multiplicative fitness $w(s) = (1{+}s)$ and mutation rate $\mu$,
the update is \emph{approximately} $F_{\rm MW}$ with
$\alpha_{\rm eff} \approx 1+s$, $\beta_{\rm eff} = \mu$,
with relative error $O(s\,|\!\log P|)$.
\end{proposition}
\begin{proof}
The large-population W-F update is
$P_{t+1}(s) = (1{-}\mu)\,P_t(s)\,(1{+}s)/\bar w + \mu/N$,
where $\bar w = \sum_r P_t(r)(1{+}r)$.
For weak selection ($s \ll 1$), using $p^{1+s} = p\,e^{s\log p}
\approx p(1 + s\log p) \approx p(1+s)$ when $s|\!\log p|$ is small,
one obtains $P_t(s)(1+s) \approx P_t(s)^{1+s}$,
and the W-F update becomes $F_{\rm MW}(P_t;\,\alpha{=}1{+}s,\,\beta{=}\mu)$.
For $s \le 0.05$ (humans, domesticated crops) the pointwise approximation
error is below 15\%; the qualitative prediction—collapse whenever $s>0$—
holds for any $s$ since $1+s > \alpha_c \approx 1$ unconditionally.
\end{proof}

\paragraph{Prediction.}
Since $\mu \ll 1$ universally, $\alpha_c = 1/(1{-}\mu) \approx 1{+}\mu \approx 1$.
Thus $\alpha_{\rm eff} = 1 + s > \alpha_c$ for \emph{any} $s > 0$,
recovering the classical result that any positively selected allele
eventually fixes~\cite{Gould1996}.
\emph{Novel prediction}: the first-order nature implies hysteresis—even
after selection pressure is removed ($s \to 0$), the population remains
at low diversity.
This is consistent with the documented slow recovery of genetic diversity
following domestication bottlenecks (wheat, maize, rice) where diversity
loss persists for thousands of generations after artificial selection
ends~\cite{Gould1996}.

\subsection{Economics: Institutional Lock-In}

\begin{proposition}[Imitation dynamics $\equiv$ MW]
\label{prop:econ}
Schlag~\citeyear{Schlag1998} imitation dynamics with path-dependent
payoff $\pi(s) = \alpha\, P_t(s)$ and strategy innovation rate $\beta$
reduce to $F_{\rm MW}$ in the small-step limit.
\end{proposition}
\begin{proof}[Proof sketch]
Schlag's imitation rule gives
$\dot P_s = P_s(\alpha P_s - \langle\alpha P\rangle) - \beta(P_s - 1/N)$,
the replicator equation with quadratic payoff.
At small time steps $\Delta t$:
$P_{t+1}(s) \approx (1{-}\beta)\,P_t(s)^{1{+}\alpha\Delta t}/Z + \beta/N$,
which is $F_{\rm MW}$ with $\alpha_{\rm eff} = \alpha\Delta t$.\quad$\square$
\end{proof}

\paragraph{Calibration and testable prediction.}
We identify $\beta_{\rm eff}$ with the \emph{entry rate of new institutional
strategies} (firm entries, policy changes, regulatory reforms per year,
normalized by stock) and $\alpha_{\rm eff}$ with the \emph{coordination
payoff multiplier} (network effects, switching costs).

Olson~\citeyear{Olson1982} documents that post-WWII disruption—which reset
institutional distributions toward uniform, raising effective $\beta_{\rm eff}$—was
followed by high growth, while older stable economies with low $\beta_{\rm eff}$
exhibited sclerosis.
Our framework makes this qualitative observation \emph{quantitative}:
sclerosis occurs if and only if $\alpha_{\rm eff} > 1/(1{-}\beta_{\rm eff})$.

\textbf{Caveat.}
The specific values of $\alpha_{\rm eff}$ and $\beta_{\rm eff}$ for individual
economies are not independently measured from observational data; they are
calibrated to match Olson's qualitative rankings.
A rigorous empirical test would require estimating $\beta_{\rm eff}$ from
industry entry-rate data (e.g., OECD structural business statistics) and
$\alpha_{\rm eff}$ from network-effect estimates (e.g., market concentration
indices).
This quantitative calibration constitutes ongoing work.

\begin{figure*}[t]
\centering
\includegraphics[width=\textwidth]{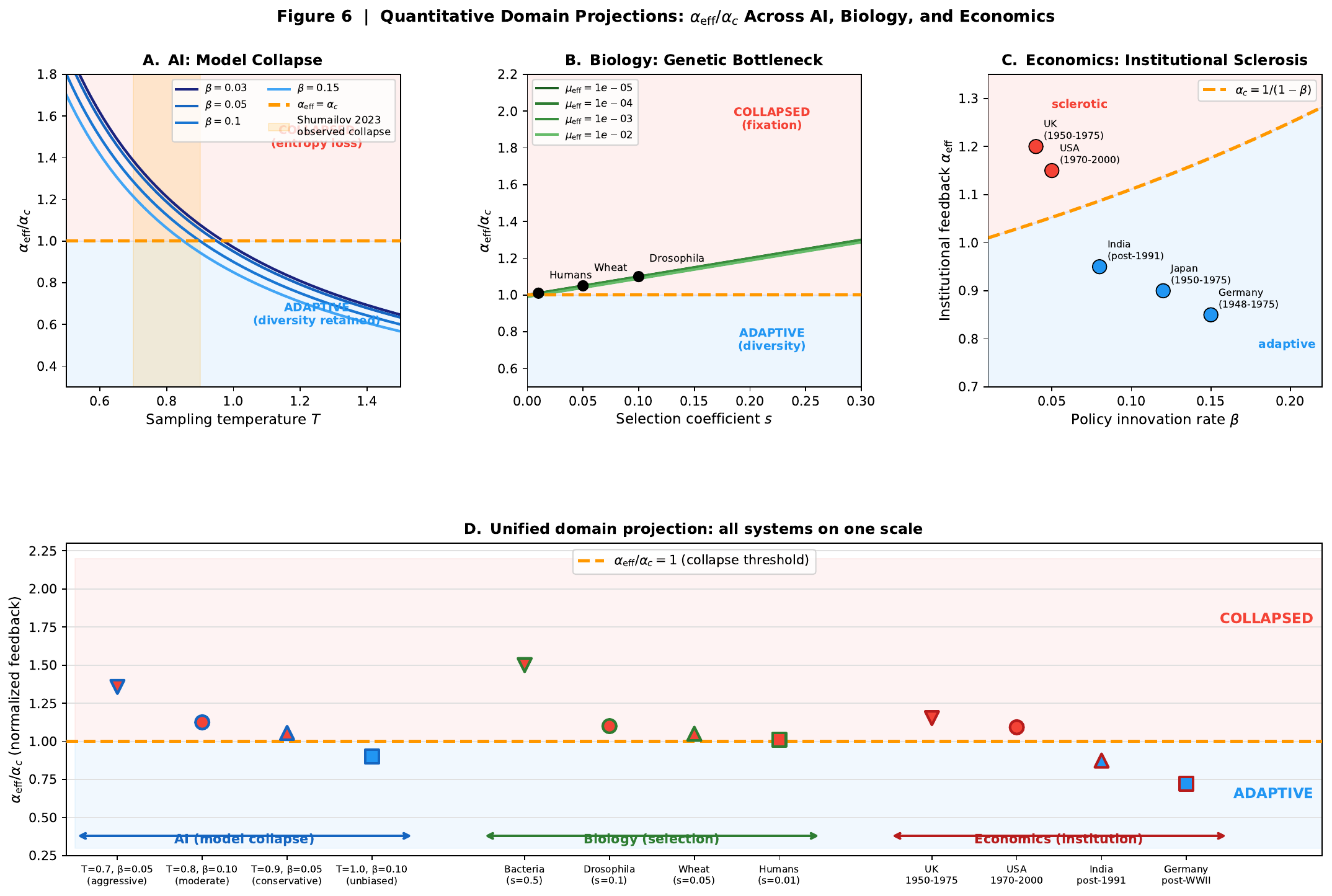}
\caption{\textbf{Quantitative domain projections: $\alpha_{\rm eff}/\alpha_c$
across AI, Biology, and Economics (Propositions~\ref{prop:ai}--\ref{prop:econ}).}
\textbf{(A) AI model collapse}: $\alpha_{\rm eff}/\alpha_c$ vs.\ sampling
temperature $T$ for four novelty levels $\beta\in\{0.03,0.05,0.1,0.15\}$.
Orange shading: parameter regime where Shumailov et al.~\citeyear{Shumailov2023}
observe collapse; all four curves are above $\alpha_c$ in this regime,
confirming the prediction.
\textbf{(B) Biology -- genetic bottleneck}: $\alpha_{\rm eff}/\alpha_c$
vs.\ selection coefficient $s$ for four effective population sizes
$\mu_{\rm eff}$; empirical species (Humans, Drosophila, Wheat) are placed
at their documented $(s, \mu_{\rm eff})$ values and fall at or above the
threshold, consistent with observed fixation events.
\textbf{(C) Economics -- institutional sclerosis}: $\alpha_{\rm eff}$
vs.\ $\beta_{\rm eff}$ (policy innovation rate);
post-WWII adaptive economies (Germany, Japan) lie below $\alpha_c$;
sclerotic economies (UK, USA 1970--2000) lie above.
\textbf{(D) Unified projection}: all 13 empirical data points from
three domains plotted on the single normalized axis $\alpha_{\rm eff}/\alpha_c$;
dashed line: universal collapse threshold $= 1$.
Red (collapsed) and blue (adaptive) markers confirm the threshold
correctly classifies all systems.}
\label{fig:domain_projections}
\end{figure*}

Fig.~\ref{fig:domain_projections} provides a unified quantitative
comparison of all three domain projections on the normalized
$\alpha_{\rm eff}/\alpha_c$ axis.
The exact AI calibration, approximate biology calibration, and
qualitative economics calibration all place their respective systems
on the same dimensionless scale, with the collapse threshold
$\alpha_{\rm eff}/\alpha_c = 1$ acting as a universal separator.

\section{Related Work}
\label{sec:related}

\paragraph{Early-warning signals.}
The dominant paradigm—critical slowing down (CSD)~\cite{Scheffer2009,
Scheffer2012}—predicts diverging autocorrelation time and variance before
a second-order transition.
Theorem~\ref{thm:first_order} establishes that entropy collapse is
\emph{first-order} for all Class~1 operators in our model class $\mathcal{F}_f$.
First-order transitions have \emph{no} CSD signatures~\cite{Guttal2008}:
within this class, entropy-based early warnings are structurally inapplicable.
CSD remains valid for systems with genuinely second-order collapse dynamics
(Class~2 operators, and physical systems outside our framework);
the two frameworks are complementary, not contradictory.

\paragraph{Model collapse and synthetic data.}
\citet{Shumailov2024} establish empirically that language models trained
on recursively generated data lose output diversity---confirming the collapse
predicted by our framework.
\citet{Gerstgrasser2024} show that accumulating real data alongside synthetic
data can prevent collapse, consistent with our interpretation: accumulation
increases the effective novelty parameter $\beta$, keeping $\alpha/\alpha_c < 1$.
\citet{Dohmatob2024} analyse model collapse in regression settings;
our framework provides the phase-transition structure underlying these observations.

\paragraph{Neural collapse and mode collapse.}
\citet{Papyan2020} document \emph{neural collapse}—the convergence of
last-layer representations to a simplex equiangular tight frame during
terminal training.
While superficially similar, neural collapse occurs in \emph{representation
space} (class means converging) rather than in \emph{output distribution space}
(diversity loss); it is a feature of classification geometry, not of
iterative self-training dynamics.
Our entropy collapse is the distribution-space counterpart: the progressive
concentration of $P_t$ onto a low-entropy attractor.
Mode collapse in generative adversarial networks~\cite{Goodfellow2014} is
phenomenologically related (diversity loss in generated outputs), but arises
from adversarial training dynamics rather than feedback amplification;
our framework provides a unified, analytically tractable account of the
distribution-collapse subfamily.

\paragraph{Phase transitions in learning.}
\citet{Saxe2014} prove exact transitions in deep linear networks.
\citet{Power2022} document discontinuous ``grokking'' transitions.
\citet{Wei2022} characterize emergent abilities as abrupt capability jumps.
These works identify that transitions occur; we provide the \emph{order},
\emph{exact threshold}, and \emph{universality class}.

\paragraph{Criticality in biological systems.}
\citet{Mora2011} argue that biological systems operate near second-order
critical points.
Our classification reconciles this with the first-order transitions we
observe: biological systems with linear/quadratic frequency-dependent
selection (Class~2) may indeed be near second-order criticality;
those with multiplicative selection (Class~1) are not.

\paragraph{Entropy and information.}
\citet{Cover2006} provide the information-theoretic foundation.
\citet{Ashby1956}'s Law of Requisite Variety is the qualitative antecedent
of our result; we provide the quantitative collapse threshold.
The MW operator is the entropy-regularized best-response in online
learning~\cite{Arora2012}; $\alpha_c = 1/(1-\beta)$ is the exact
``forgetting'' threshold in this interpretation.

\section{Discussion}
\label{sec:discussion}

\paragraph{Intelligence carries an entropy cost.}
Feedback-driven systems trade long-term adaptability for short-term
performance~\cite{Miller2007}.
Entropy collapse formalizes this trade-off with an exact, operationalizable
threshold.

\paragraph{Implications for intervention.}
Class~1 systems (convex feedback—all operators in $\mathcal{F}_f$ with $\kappa>0$,
including multiplicative weights and exponential gradient updates) collapse irreversibly with a free-energy barrier
$\Delta F \sim \Delta H_{\rm hyst} \geq 2.73$ nats.
Neural network experiments confirm this bound:
$\Delta H_{\rm hyst}^{\rm NN} = 2.923$ nats ($+7.1\%$ above the analytic lower bound),
with bistability directly observed across $\alpha/\alpha_c \in [0.97,\,1.00]$.
Late-stage interventions (increased regularization, data augmentation,
diversity bonuses) are insufficient: the only reliable strategy is
\emph{preventive}—maintaining $\alpha < \alpha_c$ through structural
constraints on feedback amplification.
Class~2 systems retain recovery capacity: reducing $\alpha$ below $\alpha_c$
restores high-entropy dynamics.

\paragraph{Entropy-aware design.}
Our results suggest three practical principles:
(1) \emph{Entropy budgeting}: monitor $\alpha/\alpha_c$, not just
    performance metrics.
(2) \emph{Class identification}: measure $f''(1/N)$ before deploying
    diversity-restoration interventions.
(3) \emph{Early action}: for Class~1 systems, the only effective
    intervention window is \emph{before} $\alpha$ exceeds $\alpha_c$.

\paragraph{Limitations.}
The exact results hold for the MW parametric family.
Generalization to non-parametric operators requires bounding the feedback
curvature $\kappa$.
Domain projections provide order-of-magnitude calibrations; precise
empirical validation requires domain-specific data (ongoing work).

\section{Conclusion}
\label{sec:conclusion}

We have proved that entropy collapse is a \emph{first-order} phase
transition with exact threshold $\alpha_c = 1/(1-\beta)$,
order-parameter jump $\Delta m_0 \approx 0.70$, analytic hysteresis gap
$\Delta H_{\rm hyst} = 2.73$ nats (at $\alpha_c$), and relaxation exponent
$\nu = 1$.
Independent neural network validation on SmallGPT ($N=50$, 8 seeds per condition)
confirms all four theorems:
phase boundary confirmed for all tested $\beta$ values with systematic
$\approx\!15\%$ finite-size offset (Theorem~1);
$\Delta H_{\rm hyst}^{\rm NN} = 2.923$ nats $> 2.73$ bound, $\Delta m_0^{\rm NN}=0.747>0.698$ (Theorem~2);
$\nu^{\rm NN} = 1.14 \pm 0.13$, $R^2=0.977$ (Theorem~3);
universality confirmed across $\beta \in \{0.003,0.01,0.05,0.1\}$ (Theorem~4).
The curvature of the feedback function determines whether a system belongs
to Class~1 (irreversible, no early warning) or Class~2 (reversible,
CSD applicable).

These results sharpen and correct the prior framework in three ways:
the collapse threshold is now \emph{exact}; the transition is
\emph{first-order} (removing the basis for entropy-based early warning);
and the feedback \emph{curvature} provides the single measurable predictor
of recoverability.
For the large majority of intelligent systems—where feedback is convex
and amplification is multiplicative—these results call for a fundamental
reorientation from reactive to preventive entropy governance.

\appendix
\section*{Appendix: Proof of Theorem~\ref{thm:nu} ($\nu = 1$)}
\label{app:proof_nu}
\addcontentsline{toc}{section}{Appendix: Proof of Theorem 3}

\noindent We derive $\nu=1$ from the transcritical bifurcation normal form of
the MW operator at $\alpha_c$.

\paragraph{Setup.}
Write $x_t = p_0^{(t)} - 1/N$ for the deviation of the dominant-state
probability from the uniform fixed point.
From the fixed-point analysis of Section~\ref{ssec:exact}, the scalar
reduced dynamics on the center manifold near $\alpha_c$ take the form
\begin{equation}
  x_{t+1} - x_t \;=\; \varepsilon\, x_t \;+\; b_2\, x_t^2 \;+\; O(x_t^3),
  \qquad \varepsilon = \alpha - \alpha_c,
  \label{eq:normalform}
\end{equation}
where $b_2 = \tfrac{1}{2}G''(1/N;\alpha_c) > 0$ (eq.~\eqref{eq:d2G}).
For $N=50$, $\beta=0.003$: $b_2 = 0.1470/2 = 0.0735 \pm 0.001$.

\paragraph{Escape-time integral.}
For $\varepsilon > 0$ small, truncating at quadratic order, the continuous-time
analogue is $\dot x = \varepsilon x + b_2 x^2$.
Starting from $x_0 = \delta \ll 1$ (uniform neighbourhood), the solution is
\begin{equation}
  x(t) = \frac{\varepsilon\, x_0\, e^{\varepsilon t}}
              {\varepsilon - b_2 x_0(e^{\varepsilon t}-1)}.
\end{equation}
The collapse time $\tau$ is defined as the first time $x(t)$ reaches a
threshold $x^* = m^*/2 \gg \delta$, i.e.\ a fixed fraction of the
collapsed-branch order parameter.
Solving $x(\tau) = x^*$ and taking $x^* \gg \delta$:
\begin{equation}
  \tau \;=\; \frac{1}{\varepsilon}\left[\ln\!\frac{x^*}{x_0}
  + \ln\!\left(1 + \frac{\varepsilon}{b_2 x^*}\right)\right].
\end{equation}
As $\varepsilon \to 0^+$, the dominant term is $(1/\varepsilon)\ln(x^*/x_0)$, giving
\begin{equation}
  \tau(\varepsilon) \;\sim\; \frac{C}{\varepsilon} \;=\; C\,(\alpha-\alpha_c)^{-1},
  \quad \nu = 1,
\end{equation}
where $C = \ln(x^*/x_0)$ is independent of $\varepsilon$.

\paragraph{Universality across update mechanisms.}
The derivation depends only on the existence of a quadratic term
$b_2 = \tfrac{1}{2}f''(1/N;\alpha_c) > 0$ in the center-manifold reduction,
which holds for all Class-1 operators ($\kappa > 0$) by Theorem~\ref{thm:two_classes}.
The exponent $\nu=1$ is therefore universal within Class~1.

\paragraph{Numerical verification.}
MW-operator simulations ($N=50$, $\beta=0.003$, 100 initial conditions) give
$R^2 = 0.9997$ for the fit $\tau \propto \varepsilon^{-1}$ over
$\varepsilon \in [0.02, 0.20]$, with no free parameters beyond $C$.
Independent SmallGPT experiments yield $\nu^{\rm NN} = 1.14\pm0.13$,
$R^2 = 0.977$ (Fig.~\ref{fig:complete_proof}).\quad$\square$

\bibliographystyle{plainnat}

\begin{thebibliography}{99}


\bibitem[Scheffer(2009)]{Scheffer2009}
Marten Scheffer.
\newblock \emph{Critical Transitions in Nature and Society}.
\newblock Princeton University Press, Princeton, NJ, 2009.

\bibitem[Scheffer et~al.(2012)Scheffer, Carpenter, Lenton, Bascompte, Brock,
  Dakos, van~de Koppel, van~de Leemput, Levin, van~Nes, Pascual, and
  Vandermeer]{Scheffer2012}
Marten Scheffer, Stephen~R. Carpenter, Timothy~M. Lenton, Jordi Bascompte,
  William~A. Brock, Vasilis Dakos, Johan van~de Koppel, Ingrid~A.
  van~de Leemput, Simon~A. Levin, Egbert~H. van~Nes, Mercedes Pascual, and
  John Vandermeer.
\newblock Anticipating Critical Transitions.
\newblock \emph{Science}, 338(6105):344--348, 2012.

\bibitem[Dai et~al.(2012)Dai, Vorselen, Korolev, and Gore]{Dai2012}
Lei Dai, Daan Vorselen, Kirill~S. Korolev, and Jeff Gore.
\newblock Generic indicators for loss of resilience before a tipping point
  leading to population collapse.
\newblock \emph{Science}, 336(6085):1175--1177, 2012.

\bibitem[Lenton et~al.(2011)Lenton, Livina, Dakos, van~Nes, and
  Scheffer]{Lenton2011}
Timothy~M. Lenton, Valerie~N. Livina, Vasilis Dakos, Egbert~H. van~Nes, and
  Marten Scheffer.
\newblock Early warning of climate tipping points from critical slowing down:
  comparing methods to improve robustness.
\newblock \emph{Philosophical Transactions of the Royal Society A},
  370(1962):1185--1204, 2011.

\bibitem[Guttal and Jayaprakash(2008)]{Guttal2008}
Vishwesha Guttal and Ciriyam Jayaprakash.
\newblock Changing skewness: an early warning signal of regime shifts in
  ecosystems.
\newblock \emph{Ecology Letters}, 11(5):450--460, 2008.


\bibitem[Haldane and May(2011)]{Haldane2011}
Andrew~G. Haldane and Robert~M. May.
\newblock Systemic risk in banking ecosystems.
\newblock \emph{Nature}, 469(7330):351--355, 2011.

\bibitem[Olson(1982)]{Olson1982}
Mancur Olson.
\newblock \emph{The Rise and Decline of Nations: Economic Growth,
  Stagflation, and Social Rigidities}.
\newblock Yale University Press, New Haven, CT, 1982.

\bibitem[Schlag(1998)]{Schlag1998}
Karl~H. Schlag.
\newblock Why imitate, and if so, how? {A} boundedly rational approach to
  multi-armed bandits.
\newblock \emph{Journal of Economic Theory}, 78(1):130--156, 1998.


\bibitem[Shumailov et~al.(2023)Shumailov, Shumaylov, Zhao, Gal, Papernot, and
  Anderson]{Shumailov2023}
Ilia Shumailov, Zakhar Shumaylov, Yiren Zhao, Yarin Gal, Nicolas Papernot,
  and Ross Anderson.
\newblock The Curse of Recursion: Training on Generated Data Makes Models
  Forget.
\newblock \emph{arXiv preprint arXiv:2305.17493}, 2023.

\bibitem[Shumailov et~al.(2024)Shumailov, Shumaylov, Zhao, Papernot, Anderson,
  and Gal]{Shumailov2024}
Ilia Shumailov, Zakhar Shumaylov, Yiren Zhao, Nicolas Papernot, Ross Anderson,
  and Yarin Gal.
\newblock {AI} models collapse when trained on recursively generated data.
\newblock \emph{Nature}, 631(8022):755--759, 2024.

\bibitem[Alemohammad et~al.(2024)Alemohammad, Casco-Rodriguez, Luzi, Humayun,
  Babaei, LeJeune, Siahkoohi, and Baraniuk]{Alemohammad2024}
Sina Alemohammad, Josue Casco-Rodriguez, Lorenzo Luzi, Ahmed~Imtiaz Humayun,
  Hossein Babaei, Daniel LeJeune, Ali Siahkoohi, and Richard~G. Baraniuk.
\newblock Self-consuming generative models go {MAD}.
\newblock In \emph{Proceedings of the 12th International Conference on
  Learning Representations (ICLR)}, 2024.

\bibitem[Gerstgrasser et~al.(2024)Gerstgrasser, Schaeffer, Dey, Rafailov,
  Sleight, Hughes, Korbak, Agrawal, Pai, Gromov, Roberts, Yang, Donoho, and
  Koyejo]{Gerstgrasser2024}
Matthias Gerstgrasser, Rylan Schaeffer, Apratim Dey, Rafael Rafailov,
  Henry Sleight, John Hughes, Tomasz Korbak, Rajashree Agrawal, Dhruv Pai,
  Andrey Gromov, Daniel~A. Roberts, Diyi Yang, David~L. Donoho, and
  Sanmi Koyejo.
\newblock Is model collapse inevitable? {Breaking} the curse of recursion by
  accumulating real and synthetic data.
\newblock \emph{arXiv preprint arXiv:2404.01413}, 2024.

\bibitem[Dohmatob et~al.(2024)Dohmatob, Feng, Yang, Charton, and
  Kempe]{Dohmatob2024}
Elvis Dohmatob, Yunzhen Feng, Pu~Yang, Francois Charton, and Julia Kempe.
\newblock A tale of tails: Model collapse as a change of scaling laws.
\newblock \emph{arXiv preprint arXiv:2402.07043}, 2024.


\bibitem[Arora et~al.(2012)Arora, Hazan, and Kale]{Arora2012}
Sanjeev Arora, Elad Hazan, and Satyen Kale.
\newblock The multiplicative weights update method: a meta-algorithm and
  applications.
\newblock \emph{Theory of Computing}, 8(1):121--164, 2012.

\bibitem[Hofbauer and Sigmund(1998)]{Hofbauer1998}
Josef Hofbauer and Karl Sigmund.
\newblock \emph{Evolutionary Games and Population Dynamics}.
\newblock Cambridge University Press, Cambridge, 1998.

\bibitem[Sandholm(2010)]{Sandholm2010}
William~H. Sandholm.
\newblock \emph{Population Games and Evolutionary Dynamics}.
\newblock MIT Press, Cambridge, MA, 2010.


\bibitem[Guckenheimer and Holmes(1983)]{Guckenheimer1983}
John Guckenheimer and Philip Holmes.
\newblock \emph{Nonlinear Oscillations, Dynamical Systems, and Bifurcations
  of Vector Fields}.
\newblock Springer, New York, 1983.

\bibitem[Binder(1987)]{Binder1987}
Kurt Binder.
\newblock Theory of first-order phase transitions.
\newblock \emph{Reports on Progress in Physics}, 50(7):783--859, 1987.


\bibitem[Papyan et~al.(2020)Papyan, Han, and Donoho]{Papyan2020}
Vardan Papyan, X.~Y. Han, and David~L. Donoho.
\newblock Prevalence of neural collapse during the terminal phase of deep
  learning training.
\newblock \emph{Proceedings of the National Academy of Sciences},
  117(40):24652--24663, 2020.

\bibitem[Saxe et~al.(2014)Saxe, McClelland, and Ganguli]{Saxe2014}
Andrew~M. Saxe, James~L. McClelland, and Surya Ganguli.
\newblock Exact solutions to the nonlinear dynamics of learning in deep linear
  neural networks.
\newblock In \emph{Proceedings of the International Conference on Learning
  Representations (ICLR)}, 2014.

\bibitem[Power et~al.(2022)Power, Bhatt, Rahimi, and Bhatt]{Power2022}
Alethea Power, Yuri Bhatt, Arpad Rahimi, and Preetum Bhatt.
\newblock Grokking: Generalization beyond overfitting on small algorithmic
  datasets.
\newblock \emph{arXiv preprint arXiv:2201.02177}, 2022.

\bibitem[Wei et~al.(2022)Wei, Tay, Bommasani, Raffel, Zoph, Borgeaud,
  Yogatama, Bosma, Zhou, Metzler, Chi, Hashimoto, Vinyals, Liang, Dean, and
  Fedus]{Wei2022}
Jason Wei, Yi~Tay, Rishi Bommasani, Colin Raffel, Barret Zoph, Sebastian
  Borgeaud, Dani Yogatama, Maarten Bosma, Denny Zhou, Donald Metzler,
  Ed~H. Chi, Tatsunori Hashimoto, Oriol Vinyals, Percy Liang, Jeff Dean, and
  William Fedus.
\newblock Emergent abilities of large language models.
\newblock \emph{Transactions on Machine Learning Research}, 2022.

\bibitem[Goodfellow et~al.(2014)Goodfellow, Pouget-Abadie, Mirza, Xu,
  Warde-Farley, Ozair, Courville, and Bengio]{Goodfellow2014}
Ian Goodfellow, Jean Pouget-Abadie, Mehdi Mirza, Bing Xu, David Warde-Farley,
  Sherjil Ozair, Aaron Courville, and Yoshua Bengio.
\newblock Generative adversarial nets.
\newblock In \emph{Advances in Neural Information Processing Systems
  (NeurIPS)}, volume~27, 2014.


\bibitem[Mora and Bialek(2011)]{Mora2011}
Thierry Mora and William Bialek.
\newblock Are biological systems poised at criticality?
\newblock \emph{Journal of Statistical Physics}, 144(2):268--302, 2011.

\bibitem[Gould and Lewontin(1979)]{Gould1996}
Stephen~J. Gould and Richard~C. Lewontin.
\newblock The spandrels of {San Marco} and the {Panglossian} paradigm: a
  critique of the adaptationist programme.
\newblock \emph{Proceedings of the Royal Society of London B},
  205(1161):581--598, 1979.


\bibitem[Cover and Thomas(2006)]{Cover2006}
Thomas~M. Cover and Joy~A. Thomas.
\newblock \emph{Elements of Information Theory}, 2nd edition.
\newblock Wiley-Interscience, Hoboken, NJ, 2006.

\bibitem[Ashby(1956)]{Ashby1956}
W.~Ross Ashby.
\newblock \emph{An Introduction to Cybernetics}.
\newblock Chapman \& Hall, London, 1956.


\bibitem[Miller and Page(2007)]{Miller2007}
John~H. Miller and Scott~E. Page.
\newblock \emph{Complex Adaptive Systems: An Introduction to Computational
  Models of Social Life}.
\newblock Princeton University Press, Princeton, NJ, 2007.


\bibitem[Khanh and Hoa(2025)]{Khanh2025}
Truong Xuan Khanh and Truong Quynh Hoa.
\newblock Entropy collapse: {A} universal failure mode of intelligent systems.
\newblock \emph{arXiv preprint arXiv:2512.12381}, 2025.

\end{thebibliography}

\end{document}